\documentclass[10pt,twocolumn,letterpaper]{article}
\usepackage[table,xcdraw,dvipsnames]{xcolor}

 \usepackage[pagenumbers]{cvpr} 

\usepackage{multirow}
\usepackage{array}

\definecolor{cvprblue}{rgb}{0.21,0.49,0.74}
\usepackage[pagebackref,breaklinks,colorlinks,allcolors=cvprblue]{hyperref}

\def\paperID{V2YAKOfzqz} 
\def\confName{CVPR}
\def\confYear{2025}

\title{Fractional Collisions: A Framework for Risk Estimation of Counterfactual Conflicts using Autonomous Driving Behavior Simulations}

\author{Sreeja Roy-Singh, Sarvesh Kolekar, Daniel P. Bonny, Kyle Foss\\
\small Nuro AI, 1290 Terra Bella Avenue, Mountain View, CA 94036\\
{\tt\small [sreeja, skolekar, dbonny, kfoss] @nuro.ai}
}

\begin{document}
\maketitle
\begin{abstract}
We present a methodology for estimating collision risk from counterfactual simulated scenarios built on sensor data from automated driving systems (ADS) or naturalistic driving databases. Two-agent conflicts are assessed by detecting and classifying conflict type, identifying the agents' roles (initiator or responder), identifying the point of reaction of the responder, and modeling their human behavioral expectations as probabilistic counterfactual trajectories. The states are used to compute velocity differentials at collision, which when combined with crash models, estimates severity of loss in terms of probabilistic injury or property damage, henceforth called \textit{fractional collisions}. The probabilistic models may also be extended to include other uncertainties associated with the simulation, features, and agents. We verify the effectiveness of the methodology in a synthetic simulation environment using reconstructed trajectories from 300+ collision and near-collision scenes sourced from VTTI's SHRP2 database and Nexar dashboard camera data. Our methodology predicted fractional collisions within 1\% of ground truth collisions. We then evaluate agent-initiated collision risk of an arbitrary ADS software release by replacing the naturalistic responder in these synthetic reconstructions with an ADS simulator and comparing the outcome to human-response outcomes. Our ADS reduced naturalistic collisions by 4x and fractional collision risk by ~62\%. The framework's utility is also demonstrated on 250k miles of proprietary, open-loop sensor data collected on ADS test vehicles, re-simulated with an arbitrary ADS software release. The ADS initiated conflicts that caused 0.4 injury-causing and 1.7 property-damaging fractional collisions, and the ADS improved collision risk in 96\% of the agent-initiated conflicts. 

\end{abstract}    
\section{Introduction}
\label{sec:intro}
The proliferation of automated driving systems (ADS) on shared public roads is expected to be supported by software releases that are evaluated to ensure safe and explainable behavior \cite{ISO8800} \cite{ISO21448}. The notion of sufficiently safe often measured by the “Absence of Unreasonable Risk” (AUR) in the United States and European Union and using “As Low As Reasonably Practicable” (ALARP) for the United Kingdom. The predictability or safety of ADS are also often compared to human-like driving to demonstrate a Positive Risk Balance (PRB), i.e. the proposition that an ADS should be at least as safe as (and ideally safer than) a human driver \cite{koopman2024}. Since sole reliance on onroad testing to evaluate ADS is prohibitively expensive \cite{Kalra2016}, simulation-based tests using sensor data collected onroad subsequently re-simulated with an ADS software test release have become popular.  Simulation tests are not only more scalable and safe, they can also target testing of failure modes known to plague ADS more than humans \cite{Cummings2024}. The most widely known application of counterfactual simulations in ADS evaluation is to assess the counterfactual outcome of on-road disengagements by safety drivers, whereby the post-disengagement ADS behavior is simulated as if the safety driver not taken over from the ADS \cite{schwall2020waymo}\cite{wang2020safety}. 

\begin{figure}
\centering
\includegraphics[width=0.8\linewidth]
{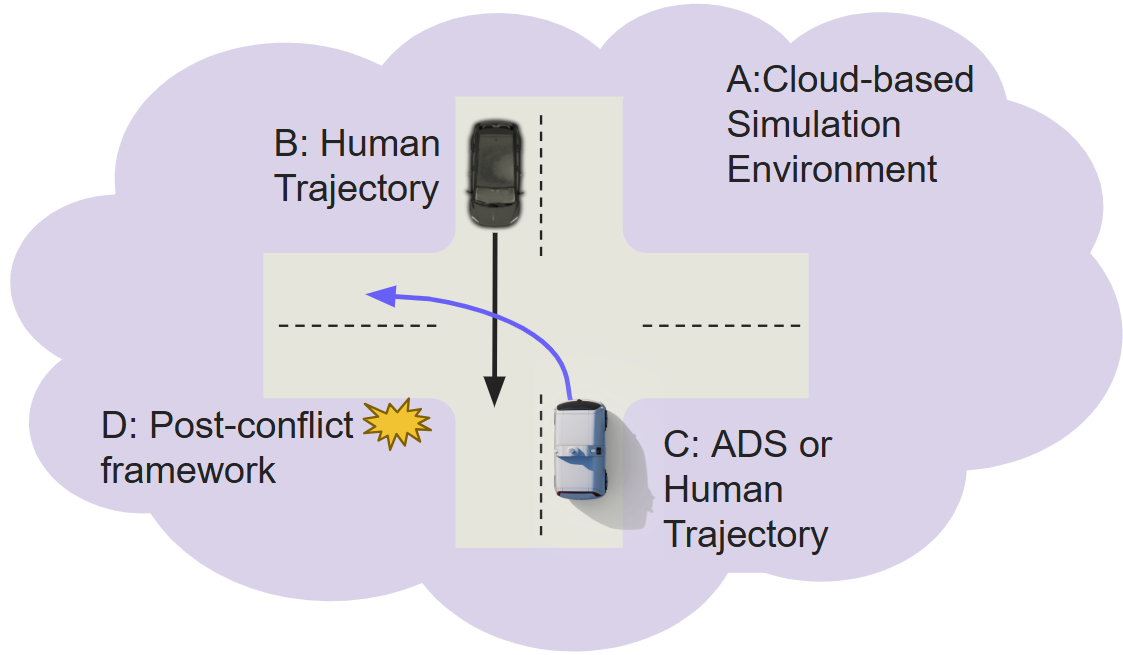}
   \caption{Simulation overview in a 2-body conflict. Agent C initiates a conflict and agent B is expected to respond to avoid a collision. For evaluating human-human performance, both trajectories are reconstructed from naturalistic databases, and for human-ADS performance, C's trajectories come from the ADS simulator and B's trajectories from either logged sensor data by an ADS test fleet or naturalistic reconstructions. This paper focuses on a framework (D) to model B's responsive behavior \textit{after} conflict onset in any sim environment (A) and any pre-conflict trajectory (B,C).}
   \label{fig:context}
\end{figure}

This paper presents a framework that provides high-fidelity estimates of collision risk in counterfactual simulations and supports both open- and closed-loop conflict data. As described in Figure \ref{fig:context}, it is applicable to any pair of agent trajectories (B and C) in a simulation environment (A) via open-loop predictions after the point of conflict initiation. While we demonstrate the framework (D) in terms of responder behavioral uncertainty only, the underlying probabilistic models may be extended to include other uncertainties associated with the ADS simulator (e.g. controller errors in C), agents (e.g. aggressiveness in B), environment (e.g. latency in A).  We also evaluate the framework and the underlying models using a novel method that uses naturalistic aggregate collision risk as ground truth. This method can be used by developers to evaluate any simulator or agent models used in the framework beyond the models described here. Finally, we show the framework's impact to ADS evaluation over large volumes of scenarios collected on our ADS test fleet or naturalistic databases. 

\section{Motivation and State of Art}
Accurate estimation of ADS-initiated collision risk requires modeling of agents that react to the ADS. Closed-loop sensor simulation on controllable simulators with reactive agents allows for dynamic evaluation of the ADS with its environment and actors \cite{yang2023waabi}\cite{hu2023wayve}, although realism of the world has been questioned \cite{Cummings2024}. Open-loop simulation replays a non-adaptive environment around the simulated ADS (Fig\ref{fig:context}-C), at significant cost savings and benefits of real-world-based conflict triggers, but requires us to explicitly model the behavior of the reacting agent (Fig\ref{fig:context}-B). The ADS can be simulated open-loop in generative synthetic environments with line following agents or using logged onroad data, where the simulated ADS can perceive the agents/objects around it and respond to them. However, the agents whose trajectories are replayed \textit{do not or cannot} respond to the ADS’s behavior, which may change from the logged version with every software release. To address this \textit{sim2real} gap, we use counterfactual simulations to model the possible behavioral reactions that a human responder may have if faced with that realistic conflict. Post-conflict counterfactuals are also relevant for hybrid simulators where some components are closed-loop (e.g. pre-conflict nominal agent behavior, sensor sim) but collision evasiveness is insufficiently modeled. 

Accurate PRB-benchmarking of agent-initiated collision risk requires modeling of agents that react to the initiator. Open-loop counterfactuals provide dense, realistic signal for agent-initiated conflicts. Such conflicts are rare on public roads because safety drivers in ADS test vehicles drive defensively, therefore simulations are the preferred test mode for coverage (or structured tests on a closed course). Pose divergence between the ADS and the logged vehicle after safety disengages motivates counterfactual situations. Using Figure\ref{fig:context} for a different use case, let's assume ADS test vehicle (C) had right of way in the onroad test, but the safety driver disengaged the ADS and yielded to B, causing B to proceed first. When the counterfactual is simulated, if the ADS (C) in the virtual environment asserts its right of way and the (non-reactive) agent is found to violate it, the scenario becomes a useful example of an agent-initiated conflict. Collision avoidance testing using accident reconstructions have benchmarked collision risk, e.g, for Waymo Driver \cite{Scanlon2021} and Baidu Apollo ADS \cite{Zhou_CAT2024}. However, they typically do not employ human behavior models because one agent's trajectory is played back exactly in the simulator while the other agent is replaced by the ADS. To address this \textit{generic benchmarking} gap, we develop probabilistic counterfactuals for human behavior to quantify their collision avoidance in place of the simulated ADS. 

Behavior modeling of reacting agents should be explainable and their outcomes comparable to the real world. Formal verification methods for scoring the risk of an ADS dynamic driving task (DDT) with respect to static obstacles and dynamic traffic agents have been well represented in the academic literature. To generate exhaustive reachable sets for all agents in a conflict, they have used Markov chains with kinematics-constrained transition functions \cite{Althoff2007}, Bayesian probabilistic occupancy grids with linear projections of kinematic parameters \cite{Ledent2019}, and data-driven learned mechanisms like higher-order control barrier functions \cite{leung2023_ddtrisk} or decision trees and random forests with natural language explanations of risk \cite{nahata2021_ddtrisk}. While these are expected to improve accuracy compared to open-loop assumptions on agent behavior such as those quantified by the Responsibility-Sensitive Safety (RSS) metrics for ADS safety \cite{shalev2017_RSS}, there are limited validation studies on real-world data with physically explainable 'why' for the human-driven 'what if' reachable sets. How and when a human driver reacts to conflict is a function of their perceived risk \cite{SUMMALA01041988}, as quantified by computational models calibrated against laboratory data \cite{kolekar2020risk}, deep learning networks with abstracted Operational Design Domain (ODD) features \cite{ping2018_LSTMrisk}, and collision avoidance difficulty using parameters like visual looming and minimal control effort needed \cite{He2024}.  Physics-based, model-in-loop counterfactual simulators to predict crash configurations \cite{leledakis2021} and simulation-based models for humans avoiding crashes are used to design more effective ADAS for nominal drivers \cite{wimmer2023} and distracted drivers \cite{bargman2024}. Human ability to navigate complex traffic situations has been modeled using future, counterfactual tracks with collision probabilities \cite{Annell2016}. Studies have compared real-world behavior to model behavior on a few conflict types and benchmark scene-sets \cite{OLLEJA2025}\cite{mattas2020}. While such sources can be used to formulate human responder models, we are not aware of any models that have been verified in terms of aggregate collision risk to establish AUR and PRB. We ensure \textit{model explainability} by formulating human-data-driven behavior models, and address the \textit{collision benchmarking} gap by embedding them into an ADS risk framework that outputs a collision metric that can be compared to observed real-world collisions.

Accurate aggregate collision risk estimation is desired within collectible data sizes. Software releases in simulation are evaluated using leading indicators of collisions: safety surrogate metrics (SSMs) and metrics indicative of defensive yet predictable driving\cite{Westhofen2023} \cite{kim2024ssmrisk}, such as compliance to the state vehicle code, comfort for the rider, and minimum disruption to other road users. Such indicators have the advantage of frequent exposure compared to collisions and therefore need smaller simulation sample sizes to evaluate ADS performance \cite{Sinha2020_ssm} \cite{alozi2023pet}. However, predictive correlations between SSMs and collisions need copious amounts of data and causality is difficult to prove. Thus, using SSMs alone for assessing absolute risk and establishing AUR/PRB is inadequate. To address this \textit{metric granularity} gap, we propose a collision risk metric that is more frequent and expected to be more statistically accurate than discrete collisions, while being more defensible in traditional safety standards than SSMs or other behavioral metrics.  
\section{Framework Description}
\label{sec:formatting}
\begin{figure}
\centering
\includegraphics[width=1.0\linewidth]
{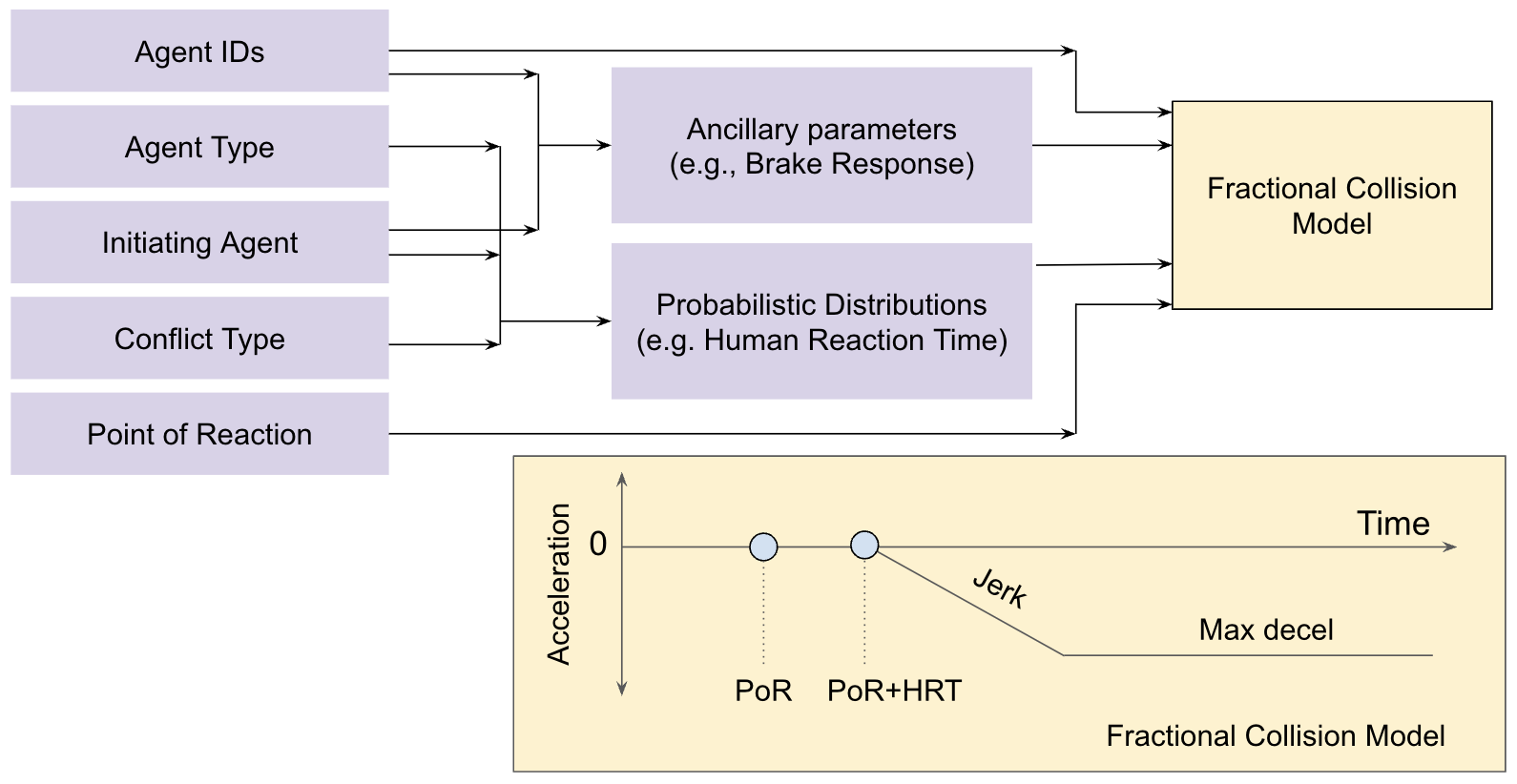}
   \caption{Summary of fractional collision workflow. Inset shows a notional reaction model for the responder in a 2-body conflict}
   \label{fig:fcs_model}
\end{figure}

The proposed collision risk framework takes as inputs trajectories of any two-agent conflict in simulation, and assesses the collision risk of that conflict using a series of steps described below (summarized in Figure \ref{fig:fcs_model}), where quality of the components are justified by referenced models that are improved upon and verified. All subsections are applicable to an open-loop simulation environment. If we use a closed-loop simulator instead, Section 3.1-3.2 can be eliminated and the remaining framework is still valid, as long as the \textit{sim2real} gap for the simulated environment and agents is sufficiently low. The framework is verified in a synthetic simulation environment using naturalistic trajectories for the agents (Section 4), and its impact on ADS development demonstrated in a synthetic and sensor replay environment using a simulated ADS as one of the agents (Section 5). 

\subsection{Classification of Conflict, Initiator, Responder}
Identified conflicts are first classified into one of 32 different types defined for our ODD. Our conflict taxonomy was initialized using NHTSA definitions (e.g  crash type variable in CRSS/FARS coding) and academic literature (e.g.,  \cite{leledakis2021}\cite{ostling2019},  \cite{kusano2023}) and extended based on internal analysis on our internal ADS test data. For example, we granularize NHTSA definition of "same trafficway, same direction, angle sideswipe" further into distinct conflict types for pullouts, lane changes, short merges, cut-ins, etc. While internal efforts are underway to automate conflict classification using machine learning (ML), the model presented here relies on heuristics for automated conflict classification with appropriate quality assurance (QA) by human triage. 

We define an \textit{initiator} as the agent in a conflict that initiates a surprising behavior that the other agent (\textit{responder}) would need to act upon to avoid propagating the conflict \cite{kusano2023} \cite{Scanlon2021}. This is not assignment of fault or blame (legal implications), and instead an automation-friendly step that ensures appropriate application of our \textit{responder} behavior models to open-loop simulation. For example, an agent phantom braking in front of a following agent on a road with no traffic lights and no threat of crossing agents ahead, makes the first agent an initiator since they started the surprising behavior.

\subsection{Behavior Models for the Responder}
The Point of Reaction (PoR) is the earliest timestamp that a responder in a simulation could have identified an impending conflict. For a human responder, the PoR is the first time the human notices conflicting or surprising behavior by the initiator. PoR also represents the timestamp at which we 'start the clock' to measure human reaction time (HRT) (Figure \ref{fig:fcs_model}-inset). Since literature-derived and proprietary HRT (Section 3.2) is very sensitive to 'start time' definitions, our models are only valid when PoR is properly defined. The definition of PoR for each conflict could be defined either from the perspective of the driver or the perspective of an omniscient third party. For the former, HRT begins when the driver has sufficient perception of the conflict, e.g., a surprise-based framework \cite{Engstrom2024}. For the latter \cite{dinakar2022_cutin}, the driver in that same conflict uses different information toward their decision, and may actually react sooner leading to the HRT distribution including negative values. While internal efforts are underway to automate PoR using ML, such as using the concept of modeling surprise, the models used in the paper rely on heuristics for automated PoR with human triage QA. 

We have built post-PoR responder models based on literature as well as trajectory data from other road users as extracted from our internal, proprietary logs of ADS driving. Our models follow the principles of Non-Impaired and Eyes on the conflict (NIEON) \cite{kusano2022} when assuming intent and ability of the responder. Inattentive agents are modeled separately as a function of known populations that cause conflicts \cite{markkula2016_lv} to account for residual risk. Kinematic reaction parameters to model the behavior of the responder after their PoR include HRT, longitudinal and lateral jerk, and steady state acceleration \cite{Bargman2017}, notionally represented in Figure \ref{fig:fcs_model}-inset. We chose HRT as the primary parameter to model spread in our behavior models due to maximum sensitivity toward severity outcomes, and longitudinal jerk and acceleration as secondary parameters. Lateral parameters were considered in a tertiary corrective step in the QA phase of the workflow due to limited data available for an automated model within acceptable uncertainties.

Reactor models are known to be sensitive to the agent type and the conflict type \cite{markkula2016_lv} hence multi-variate distributions are modeled. For example, the braking capacity of a sedan, truck, and bicycle are very different. Measurement of HRT and other kinematic parameters in the incidence of imminent danger is also sensitive to the experimental setup if naturalistic data is not readily available or has noisy dependence on naturalistic parameters. For example, the HRT for rear-end collisions\cite{markkula2016_lv} and cut-in conflicts\cite{dinakar2022_cutin} were obtained for SHRP2-based studies. Cyclist evasive parameters were found to be different when retrieved in simulation \cite{xu2017cyclist} vs. laboratory environments \cite{buchholtz2020cyclist} vs. field tests \cite{xu2017cyclist}, hence a joint distribution used. Rear-end vehicular conflicts \cite{markkula2016_lv} were parametrized differently from vehicular cut-ins \cite{dinakar2022_cutin} and motorcyclists \cite{huertas2019motorcycle}.    

\subsection{Loss Severity Mapping}
The severity of a counterfactual crash is quantified by a crash severity model (CSM). When the two colliding objects are vehicles, the model employs a momentum-based contact algorithm to estimate the velocity differentials ($delta_v$) experienced by each vehicle (Ref \cite{brach2008residual} ). If a vehicle’s mass is unknown, the mass and inertia of the vehicle are estimated based on the rectangular footprint area. The initial speeds, contact location, and relative orientation are automatically input to the model from the trajectory information. The mapping of severity vs $delta_v$ approximates the L1 and L2 severity levels ascribed by Virginia Tech Transportation Institute (VTTI) in their evaluation of SHRP2 collisions. The threshold $delta_v$ between L1 and L2 was approximated to be 6 mph. Since VTTI’s highest severity (L1) does not distinguish between injury scales, we introduced an additional level of severity (L0), thresholded at 20 mph, to categorize higher-severity crashes. 

Crash severity between a vehicle and vulnerable road users (VRUs) are evaluated using the relative resultant contact speed at time of impact. A speed below 5mph is considered an L2 event, speeds between 5 and 15 mph are considered an L1 event, and speeds above 15 mph are considered to be L0 events. Our severity gradation is roughly analogous to standardized definitions\cite{ISO26262}. IS0 26262 specifies an S1 (or higher) event to be at least 10\% probability of AIS1 (Abbreviated Injury Scale), and an S2 (or higher) event to be at least 10\% probability of AIS3. Consequently, there is a rough mapping between IS0 26262 and our severity classifications, with S0 (ISO) equivalent to L2 (proposed), S1 (ISO) equivalent to L1 (proposed) and S2 (ISO) equivalent to L0 (proposed). Alternative formulations for severity quantification verified using crash experiments and database results from Volvo \cite{waagstrom2019} may also be used.

\subsection{Risk Scoring as Fractional Collisions}
We use the severity from each counterfactual reaction in any identified conflict to estimate the collision risk of that counterfactual. We then probabilistically aggregate the severity across all counterfactual scenarios to estimate the total collision risk for that conflict as a \textit{fractional collision}. Fractional collisions can be numerically added across all conflicts in a given mileage sample size to assess the total collision risk in the form of total number of collisions. The outcome is thus easy to interpret and benchmark while the underlying models capture the uncertainties associated with various assumptions in the simulation.

\begin{figure}[t]
  \centering
   \includegraphics[width=1.0\linewidth, trim={0 6.5cm 0 1.25cm},clip]{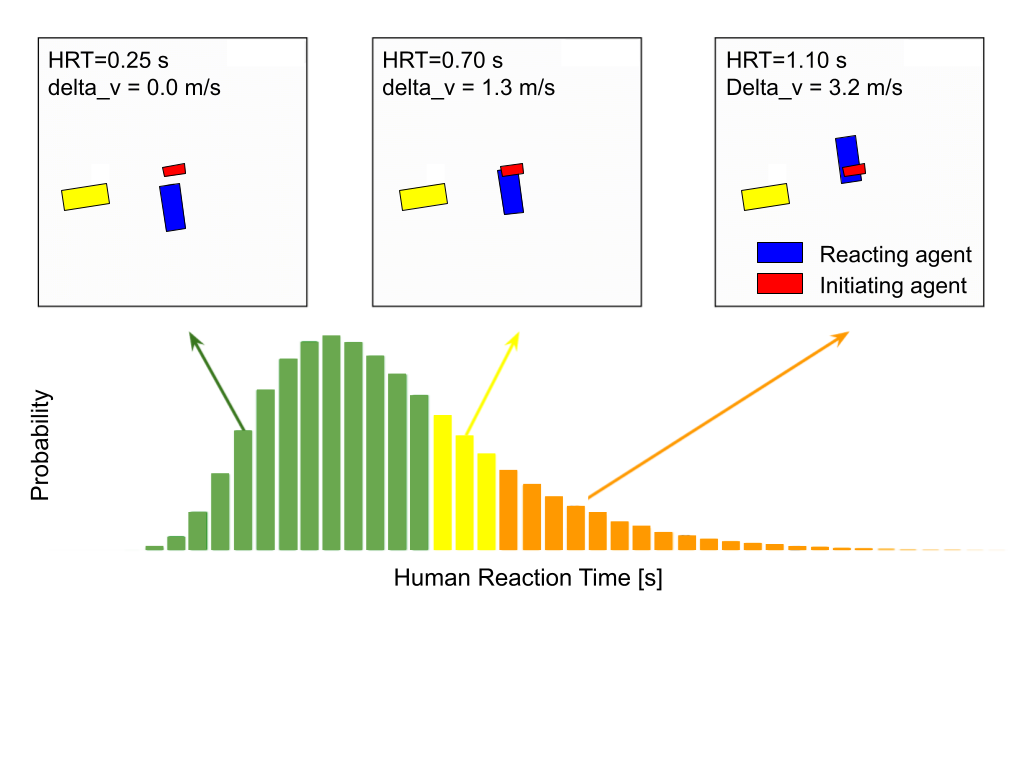}
   \caption{Example of an HRT distribution for a human responder, while holding other kinematic parameters constant in the joint probability model, used to obtain the $delta_v$ for each bin. The weighted sum across all bins is used to obtain the fractional collision score, i.e. probability of L1 (orange), L2 (yellow), no collision (green). Insets show the counterfactual behaviors that led to the arrowed bin's outcome. While a single parameter (HRT) is used for visual convenience, fractional collisions can include any number of behavioral parameters (or even extended to uncertainties associated with simulation features) whose probability mass functions are known}
   \label{fig:FCS_buildup}
\end{figure}

Figure \ref{fig:FCS_buildup} illustrates the fractional collision outcome for an example scenario across a probabilistic distribution of kinematic parameters and visualized on the HRT axis. We model the collision between a cyclist (initiator) and a vehicle (responder). After selecting an appropriate PoR (first timestamp where the cyclist is visible to the vehicle's driver and has crossed the stop line at the intersection), we simulate the brake reaction of the vehicle over its parameters. The resulting $delta_v$ for each combinatorial set of parameters is calculated and then converted to the appropriate injury severity: No collision (NC) $Lnone$ (green), property-damaging $L2$ (yellow), injury-causing $L1$ (orange), injury-causing $L0$ (red). The sum of the injury severity, weighted by the joint probability of parameters, provides the probabilistic collision value: P(NC), P(L2), P(L1), P(L2). The mass function in this example shows 20\% probability of L1, 30\% chance of L2, and 50\% chance of NC. 

Collision risk is computed as a function of the conflict initiator. Fractional collisions for ego-initiated conflicts account for the responder's probabilistic evasiveness, e.g., ADS caused 1.67 $L_x$ total collisions in spite of expected response. Note that we also compute an ADS's conflict avoidance and compliance risk using SSMs without any assumptions on responder evasiveness. For agent-initiated conflicts, we compute two loss distributions; One for ADS response, which is a single point describing the loss severity ($L_x \in [L0,L1,L2,Lnone]$c) of the collision in simulation, and another for a modeled human quantified by our probabilistic distribution of severity (like figure \ref{fig:FCS_buildup}). The difference in these distributions represents the relative risk of ADS response when compared to modeled humans for each loss severity. The aggregated risk of the ADS response across all severity can be compared to humans, e.g., ADS avoided 4.06 $L_x$ collisions compared to expected response. 

Fractional collisions can be used to evaluate the quality of ADS software. Conflicts with unreasonable collision risk are used to improve the ADS, provide feedback toward reducing the \textit{sim2real} gap, and to improve our models for risk assessment described above. Aggregate results may also be used to assess human-like driving by comparing total collision risk to naturalistic driving datasets available from NHTSA or other government sources, with known caveats for under-reporting (e.g., 55\% of all vehicles with property damage are unreported, 32\% of injuries occur in unreported crashes \cite{nhtsa2023}). While this paper focuses on losses associated with other road users and their pets, our collision risk framework can also be extended to nonhuman conflicts such as those with road debris, stray animals, and map features by mapping to different and appropriate loss severity. 
\newcommand{\subscript}[2]{$#1 _ #2$}
\section{Framework and  Fractional Collision Eval}
We assess our proposed end-to-end framework, whose components are quality assured per Section 3, against two naturalistic data sets. One data set contained agent trajectories obtained from telemetry/video data of VTTI's SHRP2 dataset, collected on voluntary subject vehicles (SV) that were retrofitted with recording instruments, and reconstructed by GCAPS (Global Center for Automotive Performance Simulation) \cite{hankey2016_shrp2}. The other data set of naturalistic agent trajectories was annotated reconstructions from dashboard camera videos provided by a commercial provider called Nexar. \cite{lavy2022_nexar}. Alternative track-annotated sources considered were APTIV's \textit{nuScenes} dataset \cite{caesar2020nuscenes} or commercially available drone or roadway camera footage of accident-prone locations such as high-speed merges or intersections. Since our workflow's verification is ideally performed on large mileage of naturalistic data across all ODDs driven randomly, static test data does not provide the coverage necessary for all conflicts. Our proprietary ADS data from public roads was considered inappropriate too because it is typically collected with an expert safety driver behind the wheel (with faster-than-average ability to perceive and react) and, when collected with the ADS in closed loop, is interrupted by ADS disengagements (which causes the part of the scene we are most interested to rarely play out naturally). 

Queried from nearly 8 million miles of SHRP2 data, 33 collisions and 497 near-collisions (NC) were found among all conflicts on surface streets involving two agents and excluding rear-end struck, animal, or inanimate object conflicts (since an evasive maneuver is not expected). Near-collisions were further restricted by those where the non-SV was at-fault since SHRP2's NC are identified based on the SV's evasive maneuver. To verify our proposed framework, we selected 17 collisions (split equally where the SV is initiator/responder) and 65 non-collisions that were qualitatively representative of the expected population distribution of possible responders for diverse scenario coverage. From an unverifiable mileage of Nexar video, 190 collisions and 44 NC were procured as potential reconstructions. The workflow described in Section 3 was applied to the selected 316 events and fractional collisions were computed as probabilities of $L_x$. 

For every scene, the fractional collision outcome is compared against the ground truth (GT) outcome, i.e. discrete $L_x$ annotated by the source (and cross-checked against our CSM in Section 3.3 applied to the conflict trajectories), as well as to a no-reaction-model (NRM) outcome, i.e. a naive control model where no deceleration is allowed after PoR. Each scene includes a GT responder should represent a datapoint on our model's joint distributions (i.e. human population), thus some scenes may have higher severity in GT than expected while others may have lower severity in GT than expected. For the same reason, we expect fewer fractional collisions than GT and NRM for scenes with collisions and more fractional collisions and NRM collisions than GT for scenes with NC. In fact, NRM may be a datapoint outside the NIEON distribution. We \textit{hypothesize} that when \textit{aggregated} across an unbiased sample of mileage across all ODDs and conflicts that a vehicle encounters, total fractional collisions should closely match GT more than NRM does. This aggregate hypothesis should also hold for a set of scenarios that represent proportionately the human population of responders. The proposed framework can be a powerful predictor of collision risk in a dataset while accounting for population variability in simulations where the agents do not respond realistically to the simulated ADS.
\subsection{Scene-by-Scene Verification}
For every scene, we implement three basic checks to ensure data quality of the reconstructions and verify that our workflow design and execution matches our intent. First, if the GT scene has an annotated collision or NC, our reconstruction of that scene through severity mapping (Section 3.3) should output as such. Second, the fractional collision outcome (Section 3.4) should not show zero probability for a severity that corresponds to the original severity. Third, the severity outcome of the NRM should be equal to or worse than GT because humans are expected to react appropriately to \textit{reduce} severity of an impending collision. 81 of 82 SHRP2 scenes were successfully reconstructed in our synthetic sim environment and passed all checks. The one exception where human reaction worsens the collision was used to improve evasive maneuver models. The Nexar-based reconstructions, owing to greater scenario volume and coverage of more conflict types and agents, surfaced need for both data quality and model improvements. Of 239 scenes, 17\% were found to be poorly reconstructed in the simulator due to sudden jerk associated with the original collision being extrapolated unrealistically for longer HRTs or NRM. Of the others, $\sim$10\% failed one of the three basic checks due to framework-intrinsic reasons such as vehicular polygons in the reconstructions being underestimated, PoR being inaccurate, and original human driver being fully nonreactive (violating NIEON).

Our probabilistic framework is more accurate in predicting the most likely outcome but also ensures full recall of worst outcomes in any scene. The most probable severity predicted by fractional collisions, i.e. the severity $L_x$ corresponding to $max[P(L_x)]$, matched the GT severity for 91\% of the scenes. The maximum fractional collision, i.e. the most severe $L_x \forall P(L_x) \geqslant 0$, matched the GT severity in only 62\% of the scenes, however was lower than GT severity in \textit{all} scenes. The most probable severity and maximum fractional collision matched GT in 51\% and 71\% Nexar scenes, respectively, and surfaced GT severity in \textit{all} scenes.

\subsection{Aggregate Verification}
SHRP2 and Nexar complement each other to provide quantitative aggregates and ODD/agent coverage with explainable trends, respectively, to verify framework effectiveness. While the SHRP2 scenes were selected such that the responder spread was representative of the human population, we could not inform selection proportions with Nexar since video content was proprietary before procurement. To verify our aggregate hypothesis and as summarized in Table\ref{tab:gcaps_table}, we compare the sum total of fractional collisions computed from the synthetic simulations of all SHRP2 and Nexar scenes to the actual GT collisions in those scenes and to the number of discrete collisions that a NRM would had in those scenes, organized by whether there was a GT collision (rows) and by severity of the collisions (columns). 
 
Compared to the 17 collision scenes from SHRP2, our framework estimated 12.6 collisions, overestimating $L0$ and $L1$ but underestimating $L2$ (shaded green). In the 65 NC, i.e. severity $Lnone$, our framework estimated 4.56 fractional collisions (pink). However, when aggregated across all scenes and severity, the total fractional collisions is 17.15, only 1\% over the actual 17 and 2x closer than using the NRM which predicts 34 collisions (blue). As expected, the NRM predicts a collision in every collision-scene, but the 17 collisions in NRM are far more skewed at higher severity (yellow) than the GT’s lower severity (green). NRM shows 9 L0 collisions while there were none in GT (also yellow).

Nexar-based aggregate results show 40\% fractional collisions compared to all GT collisions. This can be partially attributed to the disproportionately high ratio of collisions (131) to NC (37) in the procured data. In comparison, our SHRP2 selections had a C:NC ratio 17:65 and the SHRP2 database for ODD-conflicts was at 1:15. While the high C:NC ratio procured was motivated to maximize difficult scenes, Nexar’s 50x C:NC ratio compared to unbiased mileage coupled with the fact that fractional collisions will score lower than discrete collisions for any GT scene, sets the aggregate up for under-prediction. Table\ref{tab:gcaps_table} shows lower fractional collisions for every severity level (shaded green). NRM over-predicts GT as expected (shaded yellow), but by only 8\% compared to 100\% in SHRP2 (blue). To predict \textit{total} collisions accurately, the dataset must contain statistically representative NC scenarios.


\begin{table*}[t]
\fontsize{9pt}{9pt}\selectfont
\centering
\bgroup
\def\arraystretch{1.5}
\begin{tabular}{lccccccccccccc}
\multicolumn{1}{c}{} &  &  & \multicolumn{5}{c}{\textbf{SHRP2}} & \textbf{} & \multicolumn{5}{c}{\textbf{Nexar}} \\ \cline{4-8} \cline{10-14} 
\multicolumn{1}{c}{\textbf{}} & \textbf{} &  & \multicolumn{4}{c}{\textbf{Severity}} & \textbf{} &  & \multicolumn{4}{c}{\textbf{Severity}} & \textbf{} \\ \cline{4-7} \cline{10-13}
\multicolumn{1}{c}{\textbf{Post-PoR Framework}} & \textbf{GT Collision?} &  & \textbf{L0} & \textbf{L1} & \textbf{L2} & \textbf{NC} & \textbf{Total} &  & \textbf{L0} & \textbf{L1} & \textbf{L2} & \textbf{NC} & \textbf{Total} \\ \hline
 & Yes &  & \cellcolor[HTML]{CEF1CD}1.16 & \cellcolor[HTML]{CEF1CD}10.31 & \cellcolor[HTML]{CEF1CD}1.13 & 4.40 & \cellcolor[HTML]{CEF1CD}12.60 &  & \cellcolor[HTML]{CEF1CD}8.75 & \cellcolor[HTML]{CEF1CD}32.82 & \cellcolor[HTML]{CEF1CD}30.77 & 58.53 & \cellcolor[HTML]{CEF1CD}72.33 \\
 & No &  & \cellcolor[HTML]{FFE4E2}0.92 & \cellcolor[HTML]{FFE4E2}1.61 & \cellcolor[HTML]{FFE4E2}2.03 & 60.41 & \cellcolor[HTML]{FFE4E2}4.56 &  & 0.00 & 1.11 & 5.29 & 30.54 & 6.41 \\
\multirow{-3}{*}{Fractional Collisions} & All &  & 2.08 & 11.92 & 3.16 & 64.81 & \cellcolor[HTML]{C8D6FF}17.15 &  & 8.75 & 33.93 & 36.07 & 89.07 & \cellcolor[HTML]{C8D6FF}78.74 \\ \hline
 & Yes &  & \cellcolor[HTML]{CEF1CD}0 & \cellcolor[HTML]{CEF1CD}10 &\cellcolor[HTML]{CEF1CD}7 & 0 & \cellcolor[HTML]{CEF1CD}17 &  & \cellcolor[HTML]{CEF1CD}11 & \cellcolor[HTML]{CEF1CD}61 & \cellcolor[HTML]{CEF1CD}59 & 0 & \cellcolor[HTML]{CEF1CD}131 \\
 & No &  & \cellcolor[HTML]{FFE4E2}0 & \cellcolor[HTML]{FFE4E2}0 & \cellcolor[HTML]{FFE4E2}0 & 65 & \cellcolor[HTML]{FFE4E2}0 &  & 0 & 0 & 0 & 37 & 0 \\
\multirow{-3}{*}{Ground Truth (GT)} & All &  & \cellcolor[HTML]{FFFFC7}0 & 10 & 7 & 65 & \cellcolor[HTML]{C8D6FF}17 &  & \cellcolor[HTML]{FFFFC7}11 & \cellcolor[HTML]{FFFFC7}61 & \cellcolor[HTML]{FFFFC7}59 & 37 & \cellcolor[HTML]{C8D6FF}131 \\ \hline
 & Yes &  & \cellcolor[HTML]{FFFFC7}2 & \cellcolor[HTML]{FFFFC7}13 & \cellcolor[HTML]{FFFFC7}2 & 0 & \cellcolor[HTML]{FFFFC7}17 &  & 20 & 73 & 38 & 0 & 131 \\
 & No &  & 7 & 6 & 4 & 48 & 17 &  & 1 & 3 & 6 & 27 & 10 \\
\multirow{-3}{*}{No Reaction Model (NRM)} & All &  & \cellcolor[HTML]{FFFFC7}9 & 19 & 6 & 48 & \cellcolor[HTML]{C8D6FF}34 &  & \cellcolor[HTML]{FFFFC7}21 & \cellcolor[HTML]{FFFFC7}76 & \cellcolor[HTML]{FFFFC7}44 & 27 & \cellcolor[HTML]{C8D6FF}141
\end{tabular}
\egroup
\caption{Collision severity for selected SHRP2 and Nexar scenarios as a function of the framework used to model the post-PoR behavior of the responder, further split by whether there was an actual collision on-road (GT). Colors are explained in the main text narrative. }
\label{tab:gcaps_table}
\end{table*}

\section{Framework Utility in ADS Behavior Eval}

Our verified framework can be used to evaluate an arbitrary ADS software release (Fig\ref{fig:context}-C as ADS) in simulation environments (Fig\ref{fig:context}-A) to demonstrate the impact of collision risk outcomes predicted. The simulation environment can be of two types: (i) a high-volume re-simulation of logged sensor data, and (ii) a high-volume synthetic simulation of imagined, virtual environments and agents - both for Fig\ref{fig:context}-B. The ADS trajectory is the output of our proprietary simulator that plays out the ADS onboard software in a virtual environment as it 'drives' through perception inputs that are either synthetic or open-loop replayed from sensor logs. The simulator latency models are trained (and validated) using onroad and hardware-in-loop data so that each module's latency distributions are representative of the real world. Our framework (Fig\ref{fig:context}-D) enables surfacing ADS issues, estimates collision risk in ADS-initiated conflicts and collision risk eliminated by ADS in agent-initiated conflicts, and supports policy discussions on AUR and PRB. 


\begin{figure}
\centering
\includegraphics[width=1.0\linewidth]
{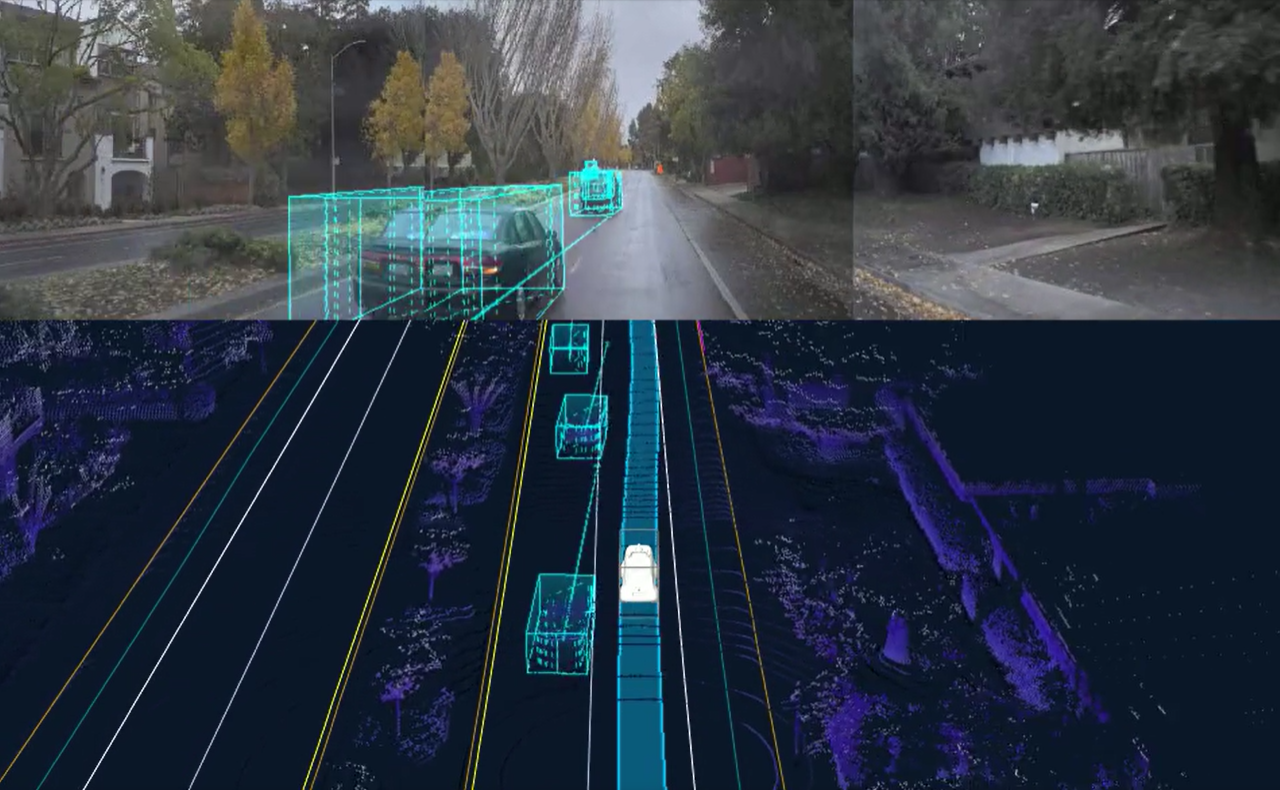}
   \caption{Example of an agent-initiated conflict with non-zero fractional collision risk in simulation. The other agent accelerates while changing lanes into the ADS', which nudges away and slows down but does not avoid collision. Screenshot marks the point of reaction of the ADS. The fractional collision results indicate that 56\% of human responders would be able to avoid this collision.}
   \label{fig:agent_collision}
\end{figure} 

\begin{figure*}[h]
  \centering
  \begin{subfigure}{0.49\textwidth}
    \includegraphics[width=\linewidth]{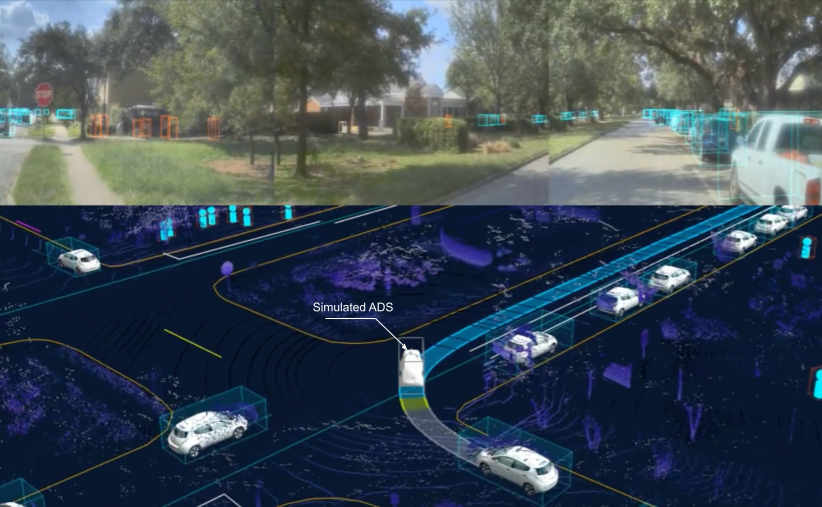} \
  \end{subfigure}
  \hfill
  \begin{subfigure}{0.49\textwidth}
    \includegraphics[width=\linewidth]{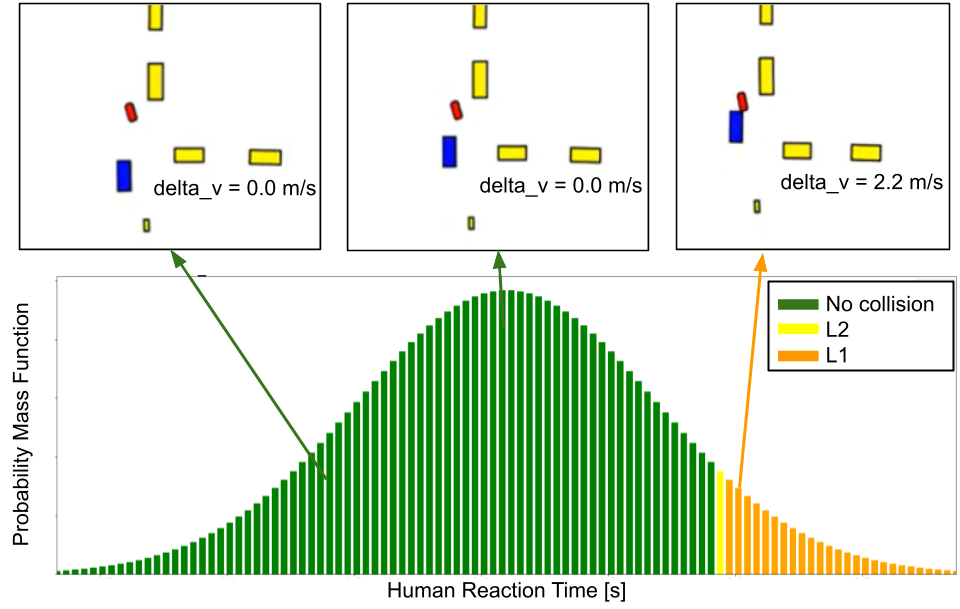}\
  \end{subfigure}
\caption{[Left] Example of an ADS-initiated conflict with non-zero fractional collision risk in simulation. The ADS unnecessarily yields during an unprotected right turn. Screenshot marks the PoR by the responder. The fractional collision results indicate that 91.5\% of human responders would be able to avoid this collision. [Right] Probability mass function associated with the HRT distribution for this conflict type, for a particular combination of other probabilistic kinematic parameters, used to compute fractional collisions}
\label{fig:ego_collision}
\end{figure*}

\subsection{Agent-Initiated Conflicts}
Replay simulation is an important mode of sourcing realistic agent-initiated conflicts because track overlaps are more frequent than in onroad testing. This is due to the pose divergence artifact described in Section 2, causing the initiator (non-ADS) to 'behave' unrealistically aggressively. Fractional collisions in such scenarios can surface agent-initiated risk and realistically quantifying PRB per scene. 376 successfully simulated agent-initiated conflicts were identified in a quarter-million mile \textit{\textbf{replay simulation}}. Of these, only 13 showed a fractional collision severity lower than the severity of the ADS's simulated outcome. The PRB gap in 3.5\% scenes is a quantitative measure of the ADS performance, and the 13 examples are opportunities for autonomy improvement because some humans could outperform the ADS (by either mitigating/lowering the severity of collision or avoiding the collision entirely).  Figure \ref{fig:agent_collision} shows an example of a conflict where the ADS is driving on a two-lane road at 35 mph. The vehicle in the left lane cuts into the lane of the ADS; due to pose divergence, this cut-in occurs approximately adjacent to the ADS. The simulated ADS vehicle nudges away from the initiator and decelerates moderately but still has an L2 collision. The modeled human shows fractional collisions of 6\% L1, 38\% L2, and 56\% no collision. Since more than half (56\%) of the humans modeled performed better than the ADS, this scene is an opportunity for autonomy improvement.

205 scenes with assured data quality were selected from the SHRP2 and Nexar sets in Section 4, reconstructed in our proprietary \textit{\textbf{synthetic simulation}} environment, and simulated with the responder replaced with our simulated ADS. The simulated ADS resulted in 26 collisions, compared to 103 collisions in the original human outcome (i.e. ADS reduced actual collisions by 4x), and 68.42 fractional collisions by the modeled human (i.e. ADS reduced expected collisions by 2.6x). In the scenes that ADS simulated to a collision, only 4 had fractional collisions better than the ADS loss. Thus, 97\% of a set of handpicked hardest scenarios sourced from tens of millions of miles of naturalistic driving confirmed that a human could not improve collision risk compared to ADS. Our framework establishes PRB for the vast majority of agent-initiated conflicts and enables the automated identification of rare agent-initiated issues to fix. 

\subsection{ADS-Initiated Conflicts}
The detection of ADS-initiated conflicts in any simulation (\textit{\textbf{replay or synthetic}}) with appropriate risk scoring is important for discovering issues and prioritizing fixes, especially to establish AUR. The vast majority of our synthetic simulations are intended to provide coverage beyond replay sim mileage for ADS-initiated conflicts. Our proposed framework can identify issues from their outcomes based on collision risk (in parallel to SSMs or other metrics). Fractional collisions increases our issue discovery signal. From the quarter-million mile log simulation, six conflicts involving the ADS and another vehicle were found with non-zero fractional collisions. These six fractional collisions can be numerically added to estimate the total, expected quantity of losses of each severity: 0 L0, 0.4 L1, 1.7 L2. Thus, our framework surfaced 6 scenarios with injury-causing risk despite no collision being likely in that mileage. 

As a concrete example of an individual scenario, Figure \ref{fig:ego_collision} shows an example conflict resulting from an unnecessary yield by the ADS while making an unprotected right turn into traffic. Cross-traffic had been yielding to a crossing school bus, providing the ADS an opportunity to make the turn; however, the ADS then unnecessarily stops after entering the lane. The responder does not react to the ADS in simulation due to pose divergence, and thus the scenario was flagged as a conflict. Our framework showed that the conflict had probabilities of 8.5\% for L1 and 91.5\% for no collision. Note that pose divergence in log replays can cause unrealistic responder behavior between PoR and HRT onset (for example, acceleration), thereby inflating risk. Since ADS should not initiate collision-risking conflicts and our goal is to discover must-fix issues, this inflation is accepted to prioritize recall.  
\section{Future Work}

Improvements expected in future work can be categorized as [1] intrinsic improvements to our proposed framework, [2] extending the framework to add uncertainties associated with the simulation environment and agents, and [3] expanding the naturalistic data set to test our updated framework on. For the first, there is work underway for various components in Section 3, especially those flagged in Section 4 (e.g. PoR). Further development and verification of agent models is required. For example, to accurately model cyclist responses, evasive maneuvers other than simple brake responses (e.g., nudging) may be considered. Simulations in Section 5.2 found six ADS-initiated conflicts involving a cyclist, where the current model appeared to overestimate the likelihood of an injury-causing collision by assuming that no responder would deviate from the logged trajectory. Upon inspection, however, a slight nudge may have been sufficient to avoid the collision. Since the SHRP2/Nexar datasets did not contain cyclist responders, more data will be needed to verify the new models. 

For the second, we will identify the most sensitive parameters in the simulation environment that affect \textit{sim2real} gap and collision risk when compared to onroad data, and model them within the proposed framework. Fractional collisions can account for some of the uncertainties associated with the tools used to evaluate them. This is especially relevant as simulators improve their capabilities in sensor modeling, neural reconstructions, world-model controlled agents, interaction sensitivity to latency, ADS controls models, etc. For the third, we will improve the quality of reconstructions, and explore sources of naturalistic data that improves representativeness of the human population and verification of the framework. 
{
    \small
    \bibliographystyle{ieeenat_fullname}
    \bibliography{main}
}


\end{document}